\documentclass{article}

\usepackage[preprint]{neurips_2021_imagenet}

\usepackage[utf8]{inputenc} 
\usepackage[T1]{fontenc}    
\usepackage{hyperref}       
\usepackage{url}            
\usepackage{booktabs}       
\usepackage{amsfonts}       
\usepackage{nicefrac}       
\usepackage{microtype}      
\usepackage{xcolor}         
\usepackage{threeparttable}
\usepackage{natbib}
\usepackage[pdftex]{graphicx}

\title{Global Attention Mechanism: Retain Information to Enhance Channel-Spatial Interactions}

\author{%
  Yichao Liu \\
  Helmholtz-Zentrum Dresden-Rossendorf\\
  Dresden, Germany\\
  \texttt{y.liu@hzdr.de} \\
  \And
  Zongru Shao \\
  Helmholtz-Zentrum Dresden-Rossendorf \\
  Dresden, Germany \\
  Center for Advanced Systems Understanding \\
  Görlitz, Germany \\
  \texttt{z.shao@hzdr.de} \\
  \And
  Nico Hoffmann \\
  Helmholtz-Zentrum Dresden-Rossendorf \\
  Dresden, Germany \\
  \texttt{n.hoffmann@hzdr.de} \\
}
\begin{document}

\maketitle

\begin{abstract}
 
A variety of attention mechanisms have been studied to improve the performance of various computer vision tasks. However, the prior methods overlooked the significance of retaining the information on both channel and spatial aspects to enhance the cross-dimension interactions. Therefore, we propose a global attention mechanism that boosts the performance of deep neural networks by reducing information reduction and magnifying the global interactive representations. We introduce 3D-permutation with multilayer-perceptron for channel attention alongside a convolutional spatial attention submodule. The evaluation of the proposed mechanism for the image classification task on CIFAR-100 and ImageNet-1K indicates that our method stably outperforms several recent attention mechanisms with both ResNet and lightweight MobileNet.

\label{Abstract}
\end{abstract}

\section{Introduction}
\label{Introduction}
Convolutional neural networks (CNNs) have been widely used in many tasks and applications in the computer vision domain (\citet{girshick2014rich, long2015fully, he2016deep, lampert2009learning}). Researchers have found that CNNs are performing well in extracting deep visual representations. With technological improvements related to CNNs, image classification on the ImageNet dataset (\citet{deng2009imagenet}) has increased from 63\% to 90\% accuracy in the past nine years (\citet{krizhevsky2012imagenet, zhai2021scaling}). Such an achievement also attributes to the complexity of the ImageNet dataset, which offers exceptional opportunities for related studies. Given the diversity and large scale of real-life scenes it covers, it has been benefitting studies for conventional image classification benchmarking, representation learning, transfer learning, etc. Particularly, it also brings challenges for the attention mechanisms.

The attention mechanisms have been improving performance in multiple applications and attracted research interests in recent years (\citet{niu2021review}). \citet{wang2017residual} used an encoder-decoder residual attention module to refine the feature maps to obtain better performance. \citet{hu2018squeeze, woo2018cbam, park2018bam} used spatial and channel attention mechanisms separately and achieved a higher accuracy. However, these mechanisms utilize visual representations from limited receptive fields due to information reduction and dimension separation. They lose global spatial-channel interactions in the process. Our research objective is to investigate attention mechanisms across the spatial-channel dimensions. We propose a ``global'' attention mechanism that reserves information to magnify the ``global'' cross-dimension interactions. Therefore, we name the proposed method Global Attention Mechanism (GAM).

\section{Related Works}

There have been several studies focusing on performance improvements of attention mechanisms for image classification tasks. Squeeze-and-Excitation Networks (SENet) (\citet{hu2018squeeze}) is the first to use channel attention and channel-wise-feature-fusion to suppress the unimportant channels. However, it is less efficient in suppressing unimportant pixels. The later-on attention mechanisms considered both spatial and channel dimensions. The convolutional block attention module (CBAM) (\citet{woo2018cbam}) places the channel and spatial attention operation sequentially, while bottleneck attention module (BAM) (\citet{park2018bam}) did it in parallel. However, both of them ignore the channel-spatial interactions and lose the cross-dimension information consequently. Considering the significance of the cross-dimension interactions, the triplet attention module (TAM) (\citet{misra2021rotate}) boosts efficiency by utilizing the attention weights between each pair of the three dimensions -- channel, spatial width, and spatial height. However, the attention operations are still applied on two of the dimensions each time instead of all three. To magnify cross-dimension interactions, we propose an attention mechanism that is capable of capturing significant features across all three dimensions.

\section{Global Attention Mechanism  (GAM)}

\begin{figure}[ht!]
  \centering
  \includegraphics[width=0.8\linewidth]{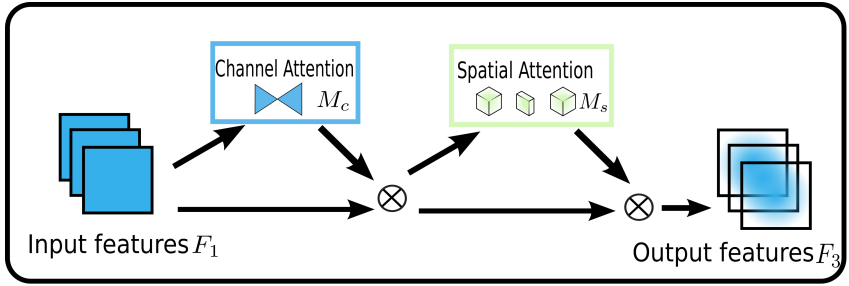}
  \caption{\textbf{The overview of GAM}}
  \label{fig:1}
\end{figure}

Our objective is to design a mechanism that reduces information reduction and magnifies global dimension-interactive features. We adopt the sequential channel-spatial attention mechanism from CBAM and redesign the submodules. The overall process is illustrated in Fig. \ref{fig:1} and formulated in Equation \ref{att:ch} and \ref{att:sp} (\citet{woo2018cbam}). Given the input feature map $\mathbf{F_1} \in \mathbb{R}^{C \times H \times W}$, the intermediate state $\mathbf{F_2}$ and the output $\mathbf{F_3}$ are defined as:

\begin{equation}
\label{att:ch}
\mathbf{F}_{2}=\mathbf{M}_{c}(\mathbf{F}_{1})\otimes \mathbf{F}_{1} 
\end{equation}
\begin{equation}
\label{att:sp}
\mathbf{F}_{3}=\mathbf{M}_{s}(\mathbf{F}_{2})\otimes \mathbf{F}_{2}
\end{equation}

where $\mathbf{M}_{c}$ and $\mathbf{M}_{s}$ are the channel and spatial attention maps, respectively; $\otimes$ denotes element-wise multiplication.

\begin{figure}[ht!]
  \centering
  \includegraphics[width=0.8\linewidth]{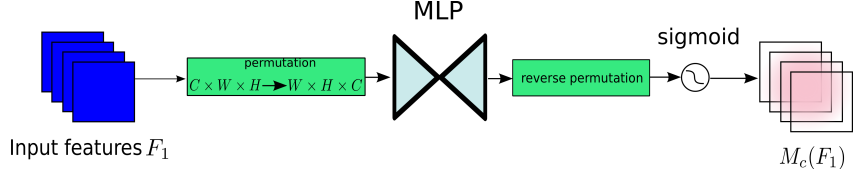}
  \caption{\textbf{Channel attention submodule.}}
  \label{fig:2}
\end{figure}

The \textbf{channel attention} submodule uses 3D permutation to retain information across three dimensions. It then magnifies cross-dimension channel-spatial dependencies with a two-layer MLP (multi-layer perceptron). (The MLP is an encoder-decoder structure with a reduction ratio $r$, same as BAM.)  The channel attention submodule is illustrated in Fig. \ref{fig:2}.

\begin{figure}[ht!]
  \centering
  \includegraphics[width=0.8\linewidth]{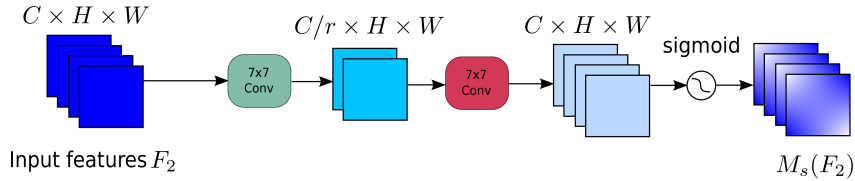}
  \caption{\textbf{Spatial attention submodule.}}
  \label{fig:3}
\end{figure}

In the \textbf{spatial attention} submodule, to focus on spatial information, we use two convolutional layers for spatial information fusion. We also use the same reduction ratio $r$ from the channel attention submodule, same as BAM. Meanwhile, max-pooling reduces the information and contributes negatively. We remove pooling to further retain the feature maps. As a result, the spatial attention module sometimes increase the number of parameters significantly. To prevent a notable increase of the parameters, we adopt group convolution with channel shuffle (\citet{zhang2018shufflenet}) in ResNet50. The spatial attention submodule without group convolution is shown in Fig. \ref{fig:3}.

\section{Experiment}
\label{Experiment}

In this section, we evaluate GAM on both CIFAR-100 (\citet{krizhevsky2009learning}) and ImageNet-1K datasets (\citet{deng2009imagenet}) with classification benchmarking and two ablation studies. We use two datasets to verify method generalization. Note that both datasets are standard for classification. ImageNet-1K has a higher impact on real-life applications.

\subsection{Classification on CIFAR-100 and ImageNet datasets}

We evaluate GAM with both ResNet (\citet{he2016deep}) and MobileNet V2 (\citet{sandler2018mobilenetv2}) as (a) they are standard architectures for image classification (b) they represent the regular and the lightweight networks respectively. We compare GAM against SE, BAM, CBAM, TAM, and Attention Branch Network (ABN) (\citet{fukui2019attention}). We re-implement the networks $\&$ mechanisms and evaluate them under the same conditions. All models are trained on four Nvidia Tesla V100 GPUs. 

For CIFAR-100, we evaluate GAM with and without group convolution (gc). We train all networks for 200 epochs with a starting learning rate of 0.1. Then, we drop the learning rate at the epochs of 60, 120, and 160. The results are shown in Table \ref{table:1}. It shows that GAM outperforms SE, BAM, and CBAM.

\begin{table}[ht!]
  \caption{Classification results on Cifar100}
  \label{table:1}
  \centering
  \begin{tabular}{lcccc}
    \toprule
    Architecture    & Parameters     & FLOPs  & Top-1 Error ($\%$) & Top-5 Error ($\%$) \\ %
    \midrule

    ResNet 50 & 23.71M  & 1.3G &  22.74 & 6.37  \\ %
    ResNet 50 + SE & 26.22M & 1.31G  & 20.29 & 5.18    \\ %
    ResNet 50 + BAM  & 24.06M & 1.33G & 19.97 &  5.03   \\ %
    ResNet 50 + CBAM & 26.24M & 1.31G   & 19.44 & 4.66   \\ %
    ResNet 50 + GAM   & 149.47M & 8.02G & \textbf{18.67} & \textbf{4.54}  \\ %
    ResNet 50 + GAM (gc*)  & 57.05M & 3.08G & 18.99 & 4.87  \\ %
    \bottomrule
  \end{tabular}
\begin{tablenotes}
     \item $\star$ gc stands for group convolution (we set its hyper-parameter as 4).
\end{tablenotes}
\end{table}

For ImageNet-1K, we pre-process the images to 224$\times$224 (\citet{he2016deep}). We include both ResNet18 and ResNet50 (\citet{he2016deep}) to verify method generalization on different network depths. For ResNet50, we include a comparison with group convolution to prevent a notable increase of the parameters. We set the starting learning rate as 0.1 and drop it for every 30 epochs. We use 90 training epochs in total. In the spatial attention submodule, we switch the first stride of the first block from 1 to 2 in order to match the size of the features. Other settings are preserved from CBAM for a fair comparison, including the use of max-pooling in the spatial attention submodule. 

MobileNet V2 is one of the most efficient lightweight models for image classification. We use the same setup of ResNet for MobileNet V2 except using an initial learning rate of 0.045 and a weight decay of $4\times10^{-5}$.

The evaluation on the ImageNet-1K is shown in Table \ref{table:2}. It shows that GAM stably enhances the performance across different neural architectures. Especially, for ResNet18, GAM outperforms ABN with fewer parameters and better efficiency.

\begin{table}[ht!]
  \caption{Classification results on ImageNet-1K}
  \label{table:2}
  \centering
  \begin{tabular}{lcccc}
  
    \toprule
    Architecture   & Parameters     & FLOPs   & Top-1 Error ($\%$) & Top-5 Error ($\%$) \\ %
    \midrule
    ResNet 18 & 11.69M  & 1.82G  & 30.91 & 11.12  \\ %
    ResNet 18 + SE   & 11.78M & 1.82G   & 30.07 & 10.59  \\ %
    ResNet 18 + BAM   & 11.71M & 1.82G  & 30.18 & 10.77   \\ %
    ResNet 18 + CBAM  & 11.78M & 1.82G   & 29.89 &  10.53  \\ %
    ResNet 18 + TAM    & 11.69M & 1.83G   & 30.0 & 10.64    \\ %
    ResNet 18 + ABN & 21.61M & 3.76G  & 29.4 & 10.34  \\ %
    ResNet 18 + GAM  & 16.04M & 2.45G  & \textbf{29.34} & \textbf{10.23}  \\
    \midrule
    ResNet 50 & 25.56M  & 4.11G & 24.81 &  7.69  \\ %
    ResNet 50 + SE   & 28.07M & 4.12G    & 23.56 & 6.82  \\ %
    ResNet 50 + BAM   & 25.92M & 4.19G  & 24.0 &  7.01 \\ %
    ResNet 50 + CBAM   & 28.09M & 4.12G   & 23.1 &  6.57   \\ %
    ResNet 50 + TAM    & 25.56M & 4.16G  & 23.29 & 6.7    \\ %
    ResNet 50 + ABN & 43.58M & 7.64G  & 23.43 & 6.92  \\ %
    ResNet 50 + GAM   & 151.32M & 24.66G  & \textbf{22.78} & \textbf{6.43}  \\ %
    ResNet 50 + GAM (gc) & 58.9M & 9.56G  & 23.01 & 6.52   \\ %
    \midrule
    MobileNet V2  & 3.51M & 0.31G    & 30.52 & 11.20 \\ %
    MobileNet V2 + SE & 3.53M & 0.32G  & 29.77 &  10.65  \\ %
    MobileNet V2 + BAM & 3.54M & 0.32G   & 29.91 & 10.80   \\ %
    MobileNet V2 + CBAM & 3.54M & 0.32G & 29.74 & 10.66    \\ %
    MobileNet V2  + GAM & 4.93M & 0.47G  & \textbf{29.31} & \textbf{10.43} \\ 
    \bottomrule
  \end{tabular}

\end{table}

\subsection{Ablation studies}

We conduct two ablation studies on ImageNet-1K with ResNet18. We first evaluate the contributions of spatial and channel attention separately. Then, we compare GAM against CBAM with and without max-pooling.

To better understand the contribution of spatial and channel attention separately, we conduct the ablation study by turning one on and the other off. For example, $ch$ indicates the spatial attention is switched off and the channel attention is on. $sp$ indicates the channel attention is turned off and the spatial attention is on. The results are shown in Table \ref{table:3}. We could observe a boost of performance on both of the on-off experiments. It indicates that both spatial and channel attentions are contributing to the performance gain. Note that their combination improves the performance with a further step.

\begin{table}[ht!]
  \caption{Ablation studies on ImageNet}
  \label{table:3}
  \centering
  \begin{tabular}{lcccc}
    \toprule
    Architecture  & Parameters     & FLOPs    & Top-1 Error ($\%$) & Top-5 Error ($\%$) \\ %
    \midrule

    ResNet 18$\dagger$ & 11.69M  & 1.82G & 30.91 & 11.12   \\ %
    ResNet 18 + GAM (sp*) &  15.95M & 2.45G & 29.61 & 10.41  \\
    ResNet 18 + GAM (ch*) & 11.78M & 1.83G & 30.25 & 10.97  \\
    ResNet 18 + GAM (ch+sp)$\dagger$ & 16.04M & 2.45G & \textbf{29.34} & \textbf{10.23}   \\
    \bottomrule
  \end{tabular}
  \begin{tablenotes}
     \item $\star$ sp stands for spatial attention only. ch stands for channel attention only.  
     \item $\dagger$ same as Table \ref{table:2}.
\end{tablenotes}
\end{table}

It is possible for max-pooling to contribute negatively in spatial attention depends on the neural architecture (e.g., ResNets). Therefore, we conduct another ablation study that compares GAM against CBAM with and without max-pooling for ResNet18. The results are shown in Table \ref{table:4}. It is observed that our method outperforms CBAM under both conditions.

\begin{table}[ht!]
  \caption{Ablation studies on ImageNet}
  \label{table:4}
  \centering
  \begin{tabular}{lcccc}
    \toprule
    Architecture  & Parameters     & FLOPs  & Top-1 Error ($\%$) & Top-5 Error ($\%$) \\
    \midrule

    ResNet 18 + CBAM$\dagger$  & 11.78M & 1.82G & 29.89   & 10.53  \\
    ResNet 18 + GAM$\dagger$  & 16.04M & 2.46G   & 29.34 & 10.23   \\
    ResNet 18 + CBAM (wmp*) & 11.78M & 1.83G  & 29.44 & 10.24   \\
    ResNet 18 + GAM (wmp*) & 16.05M & 2.47G & \textbf{28.57} & \textbf{9.83}   \\
    \bottomrule
  \end{tabular}
\begin{tablenotes}
     \item $\star$ wmp stands for without max pooling.
     \item $\dagger$ same as Table \ref{table:2}.
\end{tablenotes}
\end{table}

\section{Conclusion}
\label{Conclusion}

In this work, we proposed GAM to magnify salient cross-dimension receptive regions. Our experimental results indicate that GAM stably improves the performance for CNNs with different architectures and depths. 

CIFAR-100 and ImageNet-1K are benchmarked in our evaluation as proof of concept. They represent a scaling up with the number of classes and images. Therefore, our experiments imply that GAM is prone to data scaling capability and robustness. We consider the full ImageNet dataset serves better for applications in production. It is expensive for large-model training, especially the up-to-date top-tier solutions. Our evaluation with ResNet and MobileNet proves its feasibility on model scaling as well. We aim to investigate detailed scaling capability of GAM as the next step.

GAM obtains performance gain with an increase in the number of network parameters. In the future, we plan to investigate technologies that reduce the number of parameters for large networks, e.g., ResNet50, ResNet101, etc. Meanwhile, we also plan to explore other cross-dimension attention mechanisms that utilize parameter-reduction techniques.

\bibliographystyle{unsrtnat}
\bibliography{citation}

\end{document}